# Opportunities for artificial intelligence in advancing precision medicine

Fabian V. Filipp[*]

[*] Cancer Systems Biology, Institute of Computational Biology, Helmholtz Zentrum München, Ingolstädter Landstraße 1, D-85764 München, Germany

[*] School of Life Sciences Weihenstephan, Technical University München, Maximus-von-Imhof-Forum 3, D-85354 Freising, Germany

[*] Email: fabian.filipp@helmholtz-muenchen.de

[*] ORCID: orcid.org/0000-0001-9889-5727

**Abstract**

Machine learning (ML), deep learning (DL), and artificial intelligence (AI) are of increasing importance in biomedicine. The goal of this work is to show progress in ML in digital health, to exemplify future needs and trends, and to identify any essential prerequisites of AI and ML for precision health.

High-throughput technologies are delivering growing volumes of biomedical data, such as large-scale genome-wide sequencing assays, libraries of medical images, or drug perturbation screens of healthy, developing, and diseased tissue. Multi-omics data in biomedicine is deep and complex, offering an opportunity for data-driven insights and automated disease classification. Learning from these data will open our understanding and definition of healthy baselines and disease signatures. State-of-the-art applications of deep neural networks include digital image recognition, single cell clustering, and virtual drug screens, demonstrating breadths and power of ML in biomedicine.

Significantly, AI and systems biology have embraced big data challenges and may enable novel biotechnology-derived therapies to facilitate the implementation of precision medicine approaches.

**Introduction**

In the past decade, advances in genetic disease and precision oncology have resulted in an increased demand for predictive assays that enable the selection and stratification of patients for treatment (1). The enormous divergence of signaling and transcriptional networks mediating the cross talk between healthy, diseased, stromal and immune cells complicates the development of functionally relevant biomarkers based on a single gene or protein.

Unexpectedly, the conclusion of the human genome did not translate into a burst of new drugs. The pharmaceutical industry rather announced a declining output in terms of the number of new drugs approved despite increasing commercial efforts of drug research and development (2, 3). In contrast, machine learning (ML) as well as network and systems biology are innovating with impactful discoveries and are now starting to be seamlessly integrated into the biomedical discovery pipeline (4).

A major ambition of medical artificial intelligence (AI) lies in translating patient data to successful therapies. Machine learning models face particular challenges in biomedicine such as the size of the library to train the model, data input conversion problems, transfer, overfitting, ignorance of confounders, and many more (5-7). They may require new infrastructures, while making possibly just recently established workflows obsolete. On the other hand, deep neural network (DNN) approaches may offer distinct benefits. Such opportunities for deep learning (DL) in biomedicine include scalability, handling of extreme data heterogeneity, as well as the ability to transfer learning (8), or if wanted even the possibility not to depend on data supervision at all (9).

The goal of this work is to show progress in ML in digital health and exemplify needs, trends, and requirements for AI and ML for precision medicine. Digital image recognition, single cell analysis, and virtual screens demonstrate breadths and power of ML in biomedicine.

**Enabling synergies between artificial intelligence and digital pathology**

Advances in pattern recognition and image processing have enabled synergies between AI technology and modern pathology (10, 11). In particular, DL architectures such as deep convolutional neural networks have achieved unprecedented performance in image classification and gaming tasks (12-15). The expression "digital pathology" was coined when referring to advanced slide-scanning techniques in combination with AI-based approaches for the detection, segmentation, scoring, and diagnosis of digitized whole-slide images (16).

In pathology, quantifying and standardizing clinical outcome remains a challenge. Accurate grading, staging, classifying, and quantifying response to treatment by computer-assisted technologies are important recent initiatives (17, 18). Neural network algorithms perform well in a setting where either large amounts of input data or high quality training sets are provided. Using a digital archive of more than 100,000 clinical images of skin disease such prerequisites were fulfilled and a deep convolutional neural network was successfully trained to classify skin lesions comparable with current quality standards in pathology (19). Given such an intuitive image-based analysis, a mechanistic understanding of the convoluted layers is not necessary and the approach could be transferred to patient-based mobile phone platforms to enhance early detection and cancer prevention (20-22). In the future, specific DNN modules will replace selected steps of the traditional pathology workflow. By looking at different computational image-recognition tasks, already today, particularly strong performance of DL is already observed in segmentation tasks nuclei, epithelia or tubules, immune infiltration by lymphocyte classification, cell cycle characterization and mitosis quantification, and grading of tumors. Over time, the transition toward the digital pathology lab will lead to more accurate drug response prediction and prognosis of this underlying disease (23).

**Digital healthcare and clinical health records**

ML can learn from almost any data type, even unstructured medical text, such as patient records, medical notes, prescriptions, audio interview transcripts, or pathology and radiology reports. Future day-to-day applications will embrace ML methods to organize a growing volume of scientific literature, facilitating access and extraction of meaningful knowledge content from it (24). In the clinic, ML can harness the potential of electronic health records to accurately predict medical events (25). By implementing a ranking function in the content network, one can overcome heterogeneity of clinical or healthcare provider-specific electronic health records, inherent to the current medical practice around the world (26).

**Multi-omics integration**

A defined goal of precision medicine is to predict the best treatment strategy for the patient. Drug responses in combination with genomic, epigenomic, transcriptomic, proteomic, metabolomic profiling data provide accurate network prediction to the perturbation. Using multi-omics data, including somatic copy number alterations, somatic exome mutations, methylomes, and transcriptomes of 1000 cell lines, ML can be utilized in a modeling exercise to predict genomic features for process and drug response prediction (27). Top-performing methods exploit ML, integrate multiple profiling data sets, and enhance scoring by regression models to predict drug sensitivities (28-30). Given convolution and non-linear relationship between transcriptomic, epigenomic, and metabolic functions, future ML applications can be challenged to resolve intricate multi-omics patterns (31). Precision oncology has been showcased by implementing patient-derived cancer cell lines (32). Such bench-to-bedside models can provide real-time drug response predictions and often create of massive knowledge banks accessible to ML workup. In future, the ability to screen patient-derived avatars will inform about resistance mechanisms and facilitate evidence-based medicine, even of complex traits (33).

**Machine detection of resistance signatures**

Somatic alterations in cancer frequently escape the recognition by the endogenous immune system, creating resistance (34). Even though excellent efficacy and some complete remissions have been seen in a limited number of melanoma patients, some of whom may be regarded as cured of cancer, many malignancies show resistance or lack of response of long duration with these agents. Predicting tumor responses to immune checkpoint blockade remains a major challenge and an active field of research fueled by systems biology and AI approaches (18).

**Deciphering epigenomic networks**

Epigenomics of oncogenic networks has an ability to accurately predict regulome function, epigenomic-transcriptomic cooperation, and disease progression (35). Then again, epigenetic modifications on chromatin, DNA, and RNA are complex and often context-specific, making their mechanistic understanding challenging. Elastic net is a shrinkage method hybrid of ridge and lasso regularization (preventing over-fitting) able to handle ultra-high dimensional regression and suitable for epigenomic data (36). Using such methods, metabolic and epigenomic data have been used to establish biomarkers and to predict clocks in aging (37, 38). Enhanced by ML methods, epigenetic marks including promoter methylation are utilized as a continuous readout of transcriptional accessibility and molecular processes that guide development, tissue maintenance, disease states, and eventually aging.

Given progress in multiplex barcoding, new data challenges in the field of epigenomics are quickly at hand. Frontiers include processing and machine integration of sequencing and chromatin accessibility information derived from the transcriptome and epigenome of the same cell (39).

**Visualizing and exploring cellular heterogeneity at single cell resolution**

In single cell biology, ML and DL are frequently utilized to investigate the diversity and complexity of cell populations. In cancer, single cell methods provide a view of heterogeneity that recognizes the impact of diverse cell states and types surrounding the tumor microenvironment. Further, cancer is a dynamic and highly heterogeneous disease composed of a mix of clones characterized by distinct genotypes pushing bulk sequencing methods to their limits. Profiling of copy numbers, transcripts, or chromatin accessibility together with cluster analysis can uncover differences, even in seemingly homogenous tissues and resolve subclonal complexity. Dimensionality reduction and clustering are typical ML techniques employed to visualize single cell transcriptomics (scRNA-Seq) data. In particular, the clustering algorithm Louvain community detection is robust for high-dimensional data like scRNA-Seq matrices. The human cell atlas (40), whose primary goal is to establish, discover, and catalogue different cell populations ab initio, creates unsupervised maps, serving as a resource for subsequent disease-directed studies. In addition, it is possible to predict cycle, disease progression, and perturbation responses using deep network approaches (41-45).

Spatial transcriptomics (spRNA-Seq) combines the benefits of traditional histopathology with single cell gene expression profiling. The ability to connect the spatial organization of molecules in cells and tissues with their gene expression state enables mapping of specific disease pathology (46, 47). ML has the ability to decode molecular proximities from sequencing information and construct images of gene transcripts at sub-cellular resolution (48).

### Artificial intelligence in chemical informatics and drug discovery

Chemical informatics has an ability to predict novel drug targets, quantify ADME and toxicology, match drugs with targets and biological activities, model physicochemical properties, accelerate data mining, predict biological targets for compounds on a large scale, design new chemicals and syntheses (49), and analyze large virtual chemical spaces (50). Such a new paradigm enables medicinal chemists to process billions of molecules in virtual screens (51, 52). By tightly integrating database knowledge, AI, and lab automation it is possible to accelerate the drug discovery pipeline and select structures that can be prepared on automated systems and made available for biological testing, allowing for timely hypothesis testing and validation.

Computational analyses of drug-perturbation assays have the ability to predict the activities of the compounds on seemingly unrelated biological processes (53). ML can provide insight into drug mechanism, create correlative bridges between disjoint nodes, establish biomarkers, repurpose existing drugs, optimize drug candidates, design clinical trials, and even recruit for clinical trials. Image-based drug fingerprints were demonstrated to enable biological activity prediction for drug discovery, even when a chemical library in combination with high-content image screening was repurposed. Potential applications of predictions delivered by implemented computational models were far beyond the intended target of the original compound screen (54).

### Conclusion

Biomedical science of genomic signatures, image processing, and drug discovery rapidly adopted big data opportunities and new learning-based technologies. From traditional approaches relying on leads from nature to brute-force screening using robotics, following the introduction of several other disruptive technologies, artificial intelligence is yet another pivotal moment toward a rationalized, data driven process in healthcare and pharmaceutical industry. Machine intelligence and deep networks are changing our approach to medical bioinformatics at an unprecedented speed. As a result, the decision-making processes in precision medicine will shift from an algorithm-centric to a data-centric insight.

### Declarations

### Compliance with Ethical Standards

### Competing Interests

There is no conflict of interest.

### Human and Animal Rights and Informed Consent

This article does not contain any studies with human or animal subjects performed by any of the authors.

### Funding information

F.V.F. is grateful for the support by grants CA154887 from the National Institutes of Health, National Cancer Institute, GM115293 NIH Bridges to the Doctorate, NSF GRFP Graduate Research Fellowship Program, CRN-17-427258 by the University of California, Office of the President, Cancer Research Coordinating Committee, and the Science Alliance on Precision Medicine and Cancer Prevention by the German Federal Foreign Office, implemented by the Goethe-Institute, Washington, DC, USA, and supported by the Federation of German Industries (BDI), Berlin, Germany. This work is inspired by the curiosity and creativity of Franziska Violet and Leland Volker.

28. of network module identification across complex diseases. Nat Methods. 2019;16(9):843-52.
29. Davis S, Button-Simons K, Bensellak T, Ahsen EM, Checkley L, Foster GJ, et al. Leveraging crowdsourcing to accelerate global health solutions. Nat Biotechnol. 2019;37(8):848-50.
30. Costello JC, Heiser LM, Georgii E, Gonen M, Menden MP, Wang NJ, et al. A community effort to assess and improve drug sensitivity prediction algorithms. Nat Biotechnol. 2014;32(12):1202-12.
31. Carlberg C, Neme A. Machine learning approaches infer vitamin D signaling: Critical impact of vitamin D receptor binding within topologically associated domains. J Steroid Biochem Mol Biol. 2019;185:103-9.
32. Lee JK, Liu Z, Sa JK, Shin S, Wang J, Bordyuh M, et al. Pharmacogenomic landscape of patient-derived tumor cells informs precision oncology therapy. Nat Genet. 2018;50(10):1399-411.
33. Zeggini E, Gloyn AL, Barton AC, Wain LV. Translational genomics and precision medicine: Moving from the lab to the clinic. Science. 2019;365(6460):1409-13.
34. Zecena H, Tveit D, Wang Z, Farhat A, Panchal P, Liu J, et al. Systems biology analysis of mitogen activated protein kinase inhibitor resistance in malignant melanoma. BMC Syst Biol. 2018;12(1):33.
35. Wilson S, Filipp FV. A network of epigenomic and transcriptional cooperation encompassing an epigenomic master regulator in cancer. NPJ Syst Biol Appl. 2018;4:24.
36. Engebretsen S, Bohlin J. Statistical predictions with glmnet. Clin Epigenetics. 2019;11(1):123.
37. Ravera S, Podesta M, Sabatini F, Dagnino M, Cilloni D, Fiorini S, et al. Discrete Changes in Glucose Metabolism Define Aging. Sci Rep. 2019;9(1):10347.
38. Horvath S, Raj K. DNA methylation-based biomarkers and the epigenetic clock theory of ageing. Nat Rev Genet. 2018;19(6):371-84.
39. Chen S, Lake BB, Zhang K. High-throughput sequencing of the transcriptome and chromatin accessibility in the same cell. Nat Biotechnol. 2019.
40. Rozenblatt-Rosen O, Stubbington MJT, Regev A, Teichmann SA. The Human Cell Atlas: from vision to reality. Nature. 2017;550(7677):451-3.
41. Tian T, Wan J, Song Q, Wei Z. Clustering single-cell RNA-seq data with a model-based deep learning approach. Nature Machine Intelligence. 2019;1(4):191-8.
42. Lotfollahi M, Wolf FA, Theis FJ. scGen predicts single-cell perturbation responses. Nat Methods. 2019;16(8):715-21.
43. Wolf FA, Hamey FK, Plass M, Solana J, Dahlin JS, Gottgens B, et al. PAGA: graph abstraction reconciles clustering with trajectory inference through a topology preserving map of single cells. Genome Biol. 2019;20(1):59.
44. Wolf FA, Angerer P, Theis FJ. SCANPY: large-scale single-cell gene expression data analysis. Genome Biol. 2018;19(1):15.
45. Eulenberg P, Kohler N, Blasi T, Filby A, Carpenter AE, Rees P, et al. Reconstructing cell cycle and disease progression using deep learning. Nat Commun. 2017;8(1):463.
46. Maniatis S, Aijo T, Vickovic S, Braine C, Kang K, Mollbrink A, et al. Spatiotemporal dynamics of molecular pathology in amyotrophic lateral sclerosis. Science. 2019;364(6435):89-93.
47. Eng CL, Lawson M, Zhu Q, Dries R, Koulena N, Takei Y, et al. Transcriptome-scale super-resolved imaging in tissues by RNA seqFISH. Nature. 2019;568(7751):235-9.
48. Weinstein JA, Regev A, Zhang F. DNA Microscopy: Optics-free Spatio-genetic Imaging by a Stand-Alone Chemical Reaction. Cell. 2019;178(1):229-41 e16.
49. Segler MHS, Preuss M, Waller MP. Planning chemical syntheses with deep neural networks and symbolic AI. Nature. 2018;555(7698):604-10.
50. Ekins S, Puhl AC, Zorn KM, Lane TR, Russo DP, Klein JJ, et al. Exploiting machine learning for end-to-end drug discovery and development. Nat Mater. 2019;18(5):435-41.
51. Lo YC, Rensi SE, Torng W, Altman RB. Machine learning in chemoinformatics and drug discovery. Drug Discov Today. 2018;23(8):1538-46.
52. Chen H, Engkvist O, Wang Y, Olivecrona M, Blaschke T. The rise of deep learning in drug discovery. Drug Discov Today. 2018;23(6):1241-50.
53. Zielinski DC, Filipp FV, Bordbar A, Jensen K, Smith JW, Herrgard MJ, et al. Pharmacogenomic and clinical data link non-pharmacokinetic metabolic dysregulation to drug side effect pathogenesis. Nat Commun. 2015;6:7101.
54. Simm J, Klambauer G, Arany A, Steijaert M, Wegner JK, Gustin E, et al. Repurposing High-Throughput Image Assays Enables Biological Activity Prediction for Drug Discovery. Cell Chem Biol. 2018;25(5):611-8 e3.